%% file: amain.tex
\documentclass[journal]{IEEEtran}
\usepackage{caption}
\usepackage{color}
\usepackage{comment}
\usepackage{colortbl}  
\usepackage{xcolor}
\usepackage{epsfig}
\usepackage{graphicx}
\usepackage{bbding}
\usepackage{makecell}
\usepackage[font=footnotesize]{caption}
\usepackage{array, threeparttable}

\usepackage{amsmath}
\usepackage{amssymb}
\usepackage{comment}
\usepackage{multirow}

\usepackage{hyperref}
\hypersetup{hidelinks}

\usepackage{booktabs}
\usepackage{subfigure}
\usepackage[ruled]{algorithm2e}
\usepackage{listings}
\usepackage{color} 
\usepackage{mathtools}

\usepackage{gensymb}
\usepackage{setspace}

\usepackage{algorithmic}

\usepackage[normalem]{ulem}
\usepackage{cancel}

\begin{document}

\title{\LARGE \bf Text-Guided Multimodal Unified Industrial Anomaly Detection}

\author{Zewen Li$^{\ddagger}$, Shuo Ye$^{\ddagger}$, Zitong Yu, Weicheng Xie and Linlin Shen
\thanks{$^{\ddagger}$These authors contributed equally to this work.}
\thanks{
Corresponding authors: Zitong Yu (email: zitong.yu@ieee.org) and Linlin Shen (email: LLshen@szu.edu.cn).
}
\thanks{Z. Li, W. Xie and L. Shen are with School of Computer Science \& Software Engineering, Shenzhen University, China,
518060, and Z. Li is also with School of Computing and Information Technology, Great Bay University, Dongguan, 523000,China.}
\thanks{S. Ye and Z. Yu are with School of Computing and Information Technology, Great Bay University, Dongguan, 523000, China.}

}

\markboth{IEEE Transactions on Instrumentation and Measurement}%
{Shell \MakeLowercase{\textit{et al.}}: Bare Demo of IEEEtran.cls for IEEE Journals}

\maketitle

\begin{abstract}
Industrial anomaly detection based on RGB-3D multimodal data has emerged as a mainstream paradigm for intelligent quality inspection. However, existing unsupervised methods suffer from two critical limitations: ambiguous cross-modal alignment caused by the lack of high-level semantic guidance and insufficient geometric modeling for RGB-to-3D feature mapping. To address these issues, we propose a unified multimodal industrial anomaly detection framework guided by text semantics. The framework consists of two core modules: a Geometry-Aware Cross-Modal Mapper to preserve geometric structure during modality conversion, and an Object-Conditioned Textual Feature Adaptor to align multimodal features with semantic priors. Furthermore, we establish a unified learning paradigm for multimodal industrial anomaly detection, which breaks the “one-model-one-class” constraint and enables accurate anomaly detection across diverse classes using a single model. Extensive experiments on the MVTec 3D-AD and Eyecandies datasets demonstrate that our method achieves state-of-the-art performance in classification and localization under unsupervised settings. 
\end{abstract}


\IEEEpeerreviewmaketitle

\section{Introduction}
\label{sec1}

Industrial Anomaly Detection (IAD) is an indispensable component of intelligent quality inspection in modern manufacturing, aiming to identify unexpected defects or abnormal characteristics in industrial products with high accuracy and efficiency \cite{mvtec3d}. Due to the inherent rarity and unpredictability of anomalies in actual production, unsupervised IAD methods trained only on normal samples have become the mainstream research direction \cite{iadsurvey}. Traditional IAD methods relying solely on RGB images face inherent limitations: varying lighting conditions, subtle surface deviations, and texture color ambiguity often lead to false detection and missed detection in practical industrial scenarios \cite{mvtec3d}. The fusion of RGB images (capturing texture/color features) and 3D point cloud data (acquired by 3D sensors, providing geometric/depth information) effectively makes up for the deficiency of single-modal IAD, and has become the core research trend of IAD in recent years \cite{mvtec3d,btf,m3dm}.

\begin{figure}[t]
\centering
\includegraphics[scale=0.43]{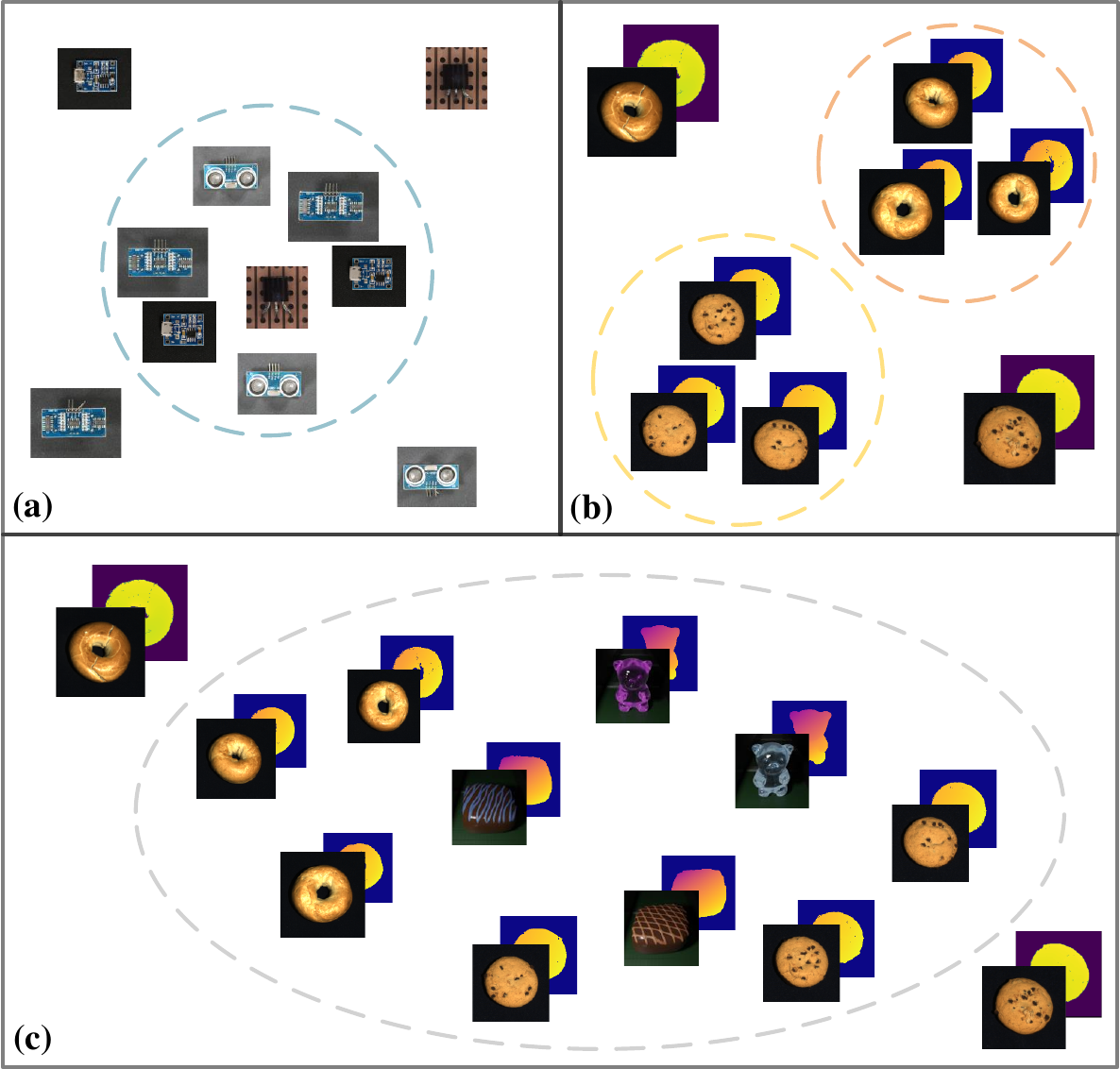}
\caption{
  Comparison of different anomaly detection settings. Normal samples within the circle, abnormal samples outside the circle. (a) Traditional Unified IAD method that train a single model to define boundaries for multiple classes using only RGB data. (b) Multi-modal IAD method adapt a one-model-one-class paradigm for multimodal inputs. (c) Multimodal unified IAD method that train a unified model to handle diverse classes across multiple modalities simultaneously.
  }
\label{fig:motivation}
\end{figure}

With the release of large-scale multimodal IAD benchmark datasets such as MVTec 3D-AD \cite{mvtec3d} and Eyecandies \cite{eyecandies}, researchers have proposed a series of unsupervised multimodal IAD methods, which can be roughly divided into memory bank-based methods \cite{btf,g2sf} and crossmodal learning-based methods \cite{m3dm,ast,cfm}. Memory bank-based methods store multimodal features of nominal samples in large memory banks and detect anomalies by measuring the distance between test features and the bank, achieving good performance but suffering from excessive memory occupancy and slow inference speed \cite{g2sf,nearestneighbor, winclip}. Crossmodal learning-based methods learn the mapping relationship between RGB and 3D features and identify anomalies by the discrepancy between predicted and actual crossmodal features, realizing lightweight inference by abandoning memory banks. However, existing multimodal IAD methods all follow a strict single-class training and inference paradigm, and three core unresolved issues jointly limit their practical industrial deployment value.

First, the lack of high-level semantic guidance leads to ambiguous crossmodal feature alignment. Existing methods only model the low-level feature interaction between RGB and 3D modalities without introducing high-level semantic information to constrain the crossmodal alignment process \cite{btf,m3dm,ast}. For industrial objects, nominal samples have definite semantic attributes (e.g., "a smooth cylindrical metal dowel", "a flat circular foam pad"), and anomalies essentially violate these intrinsic semantic attributes. Without text-based semantic prior, the crossmodal mapping model may learn spurious feature correlations between RGB and 3D modalities, leading to misalignment of semantic consistent features and reduced anomaly sensitivity. In response to the above issues, we propose to construct a joint representation learning framework between text, RGB images, and depth point clouds by introducing high-level text semantic information. By explicitly training the alignment between text semantics and multimodal structural features, this method can effectively constrain the feature mapping process of cross modalities, eliminate feature mismatches between modalities, and achieve more accurate and robust industrial anomaly detection within a unified framework.

Second, insufficient modeling of geometric characteristics causes poor RGB-to-3D mapping performance. Most existing crossmodal mapping methods implement RGB-to-3D and 3D-to-RGB mapping through lightweight MLPs, which process each pixel/point feature independently and ignore the inherent geometric spatial structure of 3D data \cite{m3dm,cfm}. RGB features are dominated by 2D RGB spatial and texture information, while 3D features contain rich geometric topology such as point adjacency and surface curvature, as well as depth information. Direct mapping between the two without geometric constraint will lead to the loss of 3D geometric characteristics in the predicted RGB-mapped 3D features, resulting in inaccurate crossmodal discrepancy calculation and blurred anomaly localization. To address these limitations, this paper proposes a GACM) designed to bridge the gap between RGB and depth modalities. Specifically, GACM employs a dual branch architecture that explicitly decouples RGB features into semantic and geometric components. By introducing a geometry-prior gating mechanism, the module adaptively suppresses non-essential texture information while amplifying structural cues such as edges and contours. This process, combined with a depth mimicking stage and a residual connection for stable feature transmission, ensures that the predicted features retain the inherent geometric topology of 3D data, thereby improving the sensitivity and localization accuracy of anomaly detection.


Third, the single-class training paradigm lacks universal adaptability for multi-class industrial scenarios. Existing multimodal IAD methods  \cite{btf,m3dm,g2sf,ast,cfm} mostly adopt the “one-model-one-class” strategy. In actual production lines involving dozens of diverse categories, deploying independent multimodal models for each category will bring huge computational and storage overhead. Although recent unified anomaly detection models represented by \cite{AA-CLIP} and \cite{adaclip} have gradually emerged to break the category-specific training limitation and pursue cross-category generalization capability, their overall detection performance still has considerable room for improvement in complex industrial environments.

To address the above three core challenges, we propose a novel multimodal IAD framework featuring text-prior alignment, geometry-aware cross-modal mapping, and universal class-unified learning. Our approach shifts the paradigm from isolated feature extraction to a unified, semantic consistent space integrating RGB, 3D and Text modalities. 
To address the aforementioned lack of semantic guidance and the resulting cross-modal misalignment, we introduce an Object-Conditioned Multimodal Alignment Module. This module shifts away from conventional low-level feature interactions by employing a Top-K Mixture-of-Experts (MoE) architecture integrated with an Object-Conditioned Textual Feature Adaptor (OCTA). By leveraging MoE to enhance text-prior embeddings, the system generates high-level semantic constraints that guide a learnable prototype via cross-attention. This mechanism effectively anchors RGB and 3D features within a unified text semantic manifold, eliminating spurious feature correlations and ensuring that multimodal representations remain consistent with the intrinsic semantic attributes of the industrial objects. Consequently, this alignment establishes a class agnostic bridge that enables robust, universal multi-class anomaly detection.
We design the GACM to mitigate the structural misalignment between RGB and 3D features. The GACM decouples RGB features into distinct semantic and geometric branches. Central to this is a geometry-prior gating mechanism that adaptively suppresses non-essential texture information such as color noise while amplifying structural cues such as edges and contours. Combined with a depth mimicking stage and residual connections, the GACM ensures that predicted features maintain high fidelity to the 3D geometric topology, resulting in significantly sharper anomaly localization.
Departing from the restrictive "one-model-one-class" paradigm, we implement a Unified Learning Strategy. By leveraging the Object-Conditioned Textual Feature Adaptor and shared cross-modal priors, our framework learns universal nominal patterns across heterogeneous industrial categories. This enables a single model to perform simultaneous anomaly detection for multiple object types, drastically reducing deployment overhead while preserving high sensitivity to fine-grained, class-specific defects.

The primary contributions of this work are summarized as follows:
\begin{itemize}
\item We introduce text semantics as high-level guidance for cross-modal alignment and construct a unified RGB-3D-Text multimodal industrial anomaly detection method. To address ambiguous cross-modal alignment and insufficient geometric modeling, we propose two core modules: the GACM to preserve geometric structure during modality conversion, and the OCTA to align multimodal features with semantic priors, which effectively eliminates ambiguous feature alignment and insufficient geometric modeling in traditional multimodal mapping.

\item With the introduction of text guidance, our method achieves excellent performance in the unified multi-class setting, breaks the “one-model-one-class” constraint, and significantly reduces deployment overhead.

\item Extensive ablation experiments and qualitative results verify the effectiveness of each module, showing that our method outperforms existing approaches in image-level classification and pixel-level defect segmentation with strong practicality and scalability.
\end{itemize}

\section{Related work}
\label{sec2}

In this section, we review the related work from four core aspects: unsupervised image anomaly detection, multimodal RGB-3D industrial anomaly detection, vision-language crossmodal alignment, and geometric feature modeling for 3D point clouds. We also highlight the research gap of universal multi-class industrial anomaly detection that has not been explored in existing works, which is the key innovation of our paper.

\subsection{Unsupervised Image Anomaly Detection}

Unsupervised IAD methods for RGB images are the foundation of multimodal IAD, and can be divided into two main categories: image reconstruction-based methods and feature embedding-based methods.

Reconstruction-based methods learn to reconstruct nominal samples using deep neural networks such as auto-encoders \cite{autoencoder1,autoencoder2}, inpainting models \cite{inpainting}, and diffusion models \cite{anoddpm}. At test time, anomalies are detected by measuring the reconstruction error between the input and reconstructed images, as the model cannot accurately reconstruct anomalous samples. Bergmann et al. \cite{autoencoder1} improved unsupervised defect segmentation by applying structural similarity to auto-encoders, and Wyatt et al. \cite{anoddpm} proposed AnoDDPM for anomaly detection using denoising diffusion probabilistic models with simplex noise. However, these methods suffer from over-smoothing of reconstructed images and low sensitivity to subtle anomalies, and the single-class training model cannot generalize to other object categories.

Feature embedding-based methods focus on modeling the distribution of nominal features in the deep feature space \cite{g2sf,uninformed,padim,patchcore,LSFA,bridgenet}, and can be further divided into memory bank-based methods and feature mapping-based methods. Memory bank-based methods \cite{uninformed,patchcore} use frozen pretrained feature extractors such as ViT and ResNet to extract high-level features of nominal samples and store them in memory banks; at inference time, anomalies are identified by comparing the distance between test features and the memory bank such as PatchCore \cite{patchcore} and G2SF \cite{g2sf}. These methods achieve remarkable performance but have high memory overhead and slow inference speed. Feature mapping-based methods such as EasyNet \cite{easynet} and CFM \cite{cfm} have advanced visual representation and anomaly detection. EasyNet constructs a lightweight feature mapping and reconstruction paradigm to efficiently fuse multi-modal features and identify anomalies with high efficiency. CFM performs adaptive masking and weighting on convolutional feature maps, which suppresses interference and enhances discriminative feature learning, thus boosting the performance of feature matching and anomaly localization.
Notably, all these methods are trained and evaluated on a single-class setting, and their feature learning process is highly class-specific, making them unable to adapt to multi-class industrial inspection scenarios.

Most of the approaches in earlier aforementioned studies follow a one-model-one-class training paradigm, which not only incurs substantial computational and memory overhead but also restricts the generalization ability of models across cross-scenario tasks. In the research field of Unified Industrial Anomaly Detection, existing methods predominantly focus on the single RGB modality.
To break the limitation of the one-model-one-class paradigm, UniAD \cite{uniad} first introduces a unified architecture capable of processing multiple categories simultaneously. By reconstructing feature embeddings and introducing a neighborhood suppression mechanism, it effectively mitigates the identity mapping problem that commonly arises during unified training. Subsequently, MoEAD \cite{moead} further advances this framework by incorporating the Mixture-of-Experts concept. It employs multiple dedicated expert networks to model the distinct feature distributions of different categories, enabling the model to preserve the convenience of unified training while more accurately capturing the highly diverse texture characteristics across various industrial products.
AnomalyGPT \cite{anomalygpt} integrates industrial images with large language models. By comparing normal and anomalous image patches, the model achieves not only pixel-level anomaly localization but also interpretable reasoning about anomaly causes through conversational interaction. This marks a paradigm shift in industrial inspection from a pure classification task toward a cognitive task. Building upon this, IAD-GPT \cite{iadgpt} leverages prompt engineering to further enhance the industrial anomaly detection capabilities of AnomalyGPT, enabling the model to simultaneously perform anomaly detection, segmentation, and anomaly class inference.

\subsection{Multimodal RGB-3D Industrial Anomaly Detection}

With the development of 3D sensing technology, multimodal RGB-3D IAD has become a research hotspot, and the release of MVTec 3D-AD \cite{mvtec3d} and Eyecandies \cite{eyecandies} datasets has greatly promoted the progress of this field. Existing multimodal RGB-3D IAD methods can be classified into memory bank-based methods, Crossmodal fusion-based methods, and all of them strictly follow the single-class training and inference paradigm, which is the most critical limitation for their industrial application \cite{mvtec3d, btf, m3dm, g2sf, ast,cfm}.

Memory bank-based methods \cite{btf,m3dm,g2sf} extend the single-modal memory bank \cite{patchcore,pr_memorybank1,pr_memorybank2} idea to multimodal scenarios by storing fused RGB-3D features in memory banks. BTF \cite{btf} added handcrafted 3D point cloud features (e.g., FPFH \cite{FPFH}) to frozen RGB features and stored them in a memory bank, achieving the first effective fusion of RGB and 3D for multimodal IAD. M3DM \cite{m3dm} further improved BTF by using Transformer-based foundation models such as DINO ViT \cite{DINO} or Point-MAE \cite{pointmae} to extract rich RGB and 3D features, and proposed a learned fusion function to generate multimodal features for memory bank storage. Although these methods achieved SOTA performance at that time, they rely on large memory banks, resulting in excessive memory occupancy and slow inference speed. More importantly, the memory bank of these methods is built on nominal samples of a single class, and the feature distribution is highly class-specific, making it impossible to expand to multi-class scenarios. G2SF \cite{g2sf} enhances the memory bank paradigm by utilizing 3D geometric cues to dynamically weight and calibrate the anomaly scores from both RGB and depth modalities, effectively resolving cross-modal decision inconsistencies.

Crossmodal fusion-based methods abandon memory banks and learn the intrinsic mapping relationship between RGB and 3D features to detect anomalies, realizing lightweight and efficient inference. 
AST \cite{ast} proposed an asymmetric teacher-student network, which takes 3D data as an additional input channel of the 2D network, but it ignores the spatial structure of 3D data and only uses it as auxiliary information, leading to inferior performance compared with CFM. CFM \cite{cfm} addresses the modality gap by learning a direct bidirectional mapping between 2D texture and 3D geometry features; it identifies anomalies by measuring the prediction discrepancy.

In addition, a small number of methods attempt to improve multimodal IAD by feature enhancement, like point cloud interpolation or feature upsampling, but they do not touch the core problems of semantic and geometric constraint in crossmodal alignment and the single-class training paradigm, and their performance improvement is limited.

\subsection{Vision-Language Crossmodal Alignment}

Vision-language crossmodal alignment aims to establish a unified semantic space for visual and text modalities, and has achieved great success in general computer vision tasks \cite{DINO,dinov2,clip}, e.g. image captioning, visual question answering and zero-shot classification. The core idea of these methods is to learn a crossmodal feature space where the embeddings of visually similar images and semantically consistent texts are close to each other \cite{dinov2}, which provides a natural solution for semantic guidance in unified IAD.

Mainstream vision-language alignment methods are based on pretrained foundation models trained on large scale image-text pairs, such as CLIP \cite{clip} and Dinov2 \cite{dinov2}. These models can map images and texts into a shared semantic space with strong crossmodal alignment ability and good generalization performance across different object categories. In industrial computer vision, vision-language alignment has been applied to IAD tasks\cite{winclip, clipad,dycclip,generalizingclip,clip_fewshot}.Addressing the limitation of CLIP in capturing local subtle defects, WinCLIP \cite{winclip} proposes a window-based analysis paradigm. By extracting patch features at multiple scales and integrating elaborately designed prompt ensembles, it achieves superior zero-shot and few-shot anomaly localization performance.
CLIP-AD \cite{clipad} and other enhanced methods \cite{anomalygpt,iadgpt} further extend this paradigm. CLIP-AD introduces a staged alignment strategy or a dual path architecture, these approaches alleviate the anomaly insensitivity issue of vanilla CLIP global features in industrial scenarios, enabling more precise spatially consistent mapping between visual features and anomaly semantics.

The key challenge of applying vision-language alignment to multimodal IAD is how to combine text prior with RGB-3D crossmodal feature learning and use text semantic information to guide both single-class and multi-class anomaly detection. Different from general vision-language tasks, IAD focuses on the nominal attribute constraints of objects: text prompts describe the essential attributes of nominal objects, and anomalies are the violation of these attributes. Our work integrates pretrained vision-language models into the RGB-3D multimodal IAD framework, constructs a RGB-3D-Text semantic alignment module, and uses text prior to constrain the crossmodal feature mapping process and establish a class agnostic semantic bridge for multiclass learning.

\subsection{Geometric Feature Modeling for 3D Point Clouds}

Modeling the geometric features of 3D point clouds is a basic research direction of 3D computer vision, and the extracted geometric features are widely used in point cloud classification, segmentation and registration \cite{FPFH, pointmae,pointnet++}. The effective modeling of 3D geometric features is the key to improving the accuracy of RGB-to-3D crossmodal mapping in multimodal IAD.

Handcrafted geometric feature descriptors are the classic representation of 3D point cloud geometric features: FPFH \cite{FPFH} extracts local geometric features of point clouds by calculating the relative position and normal vector of neighboring points, and is widely used in 3D point cloud registration; Point Feature Histograms (PFH) \cite{PFH} further improves the descriptive ability of local geometric features but with higher computational complexity. Deep learning-based methods extract high-level geometric features of point clouds through hierarchical feature learning: PointNet++ \cite{pointnet++} learns hierarchical point cloud features by sampling and grouping local point neighborhoods, capturing the global and local geometric structure of point clouds; Point-MAE \cite{pointmae} uses masked auto-encoding to pretrain point cloud transformers on large-scale 3D datasets, learning robust geometric feature representations for 3D point clouds.

\begin{figure*}[t]
\centering
\includegraphics[scale=0.6]{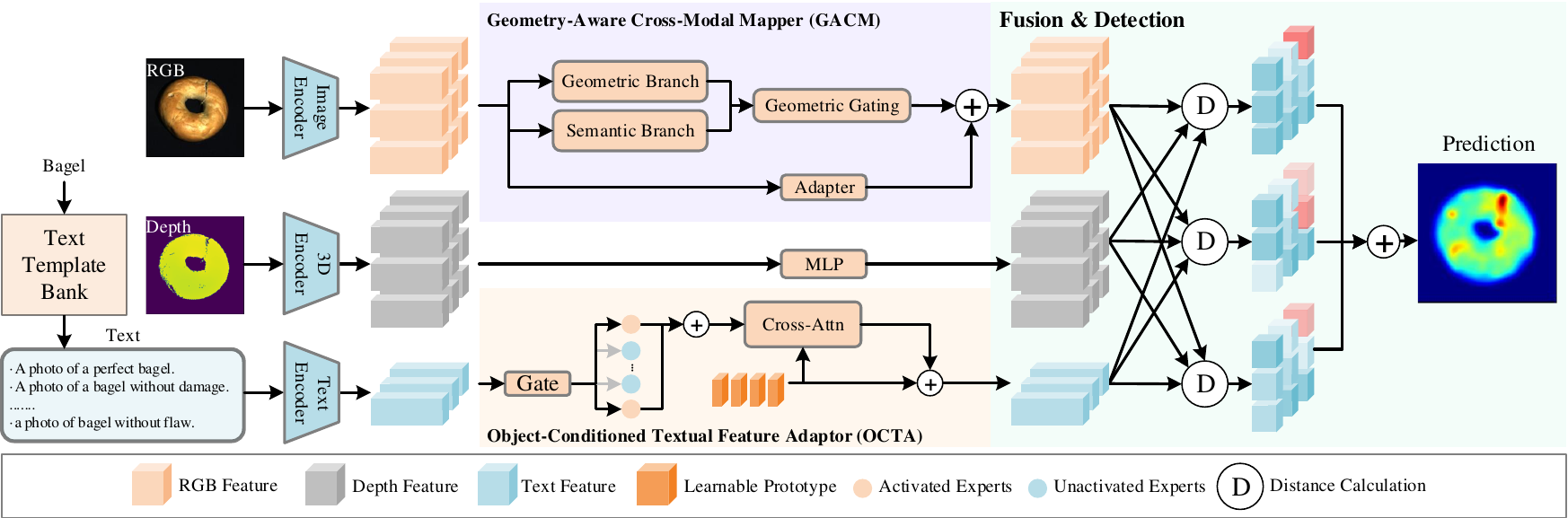}
\caption{
  Overview of our method. The framework consists of three main stages. (1) Multi-modal Feature Extraction. RGB images, depth maps, and text are processed through an Image Encoder, 3D Encoder, and Text Encoder, respectively. (2) Feature Mapping. Including modules such as GACM and OCTA to map features into different modalities. (3) Fusion $\&$ Detection. The processed multi-modal features are compared through Distance Calculation ($D$) to generate the final anomaly score map and the localized Prediction.
  }

\label{fig:model}
\end{figure*}

In multimodal RGB-3D IAD, existing methods \cite{btf,m3dm,cfm} only use deep learning-based 3D feature extractors to extract general 3D features, but do not explicitly model and fuse the handcrafted and deep geometric characteristics of 3D point clouds into the RGB-to-3D mapping process. This leads to the loss of fine-grained geometric information in the mapping process and reduced accuracy of predicted 3D features. Our work proposes a Geometry-Aware Module that fuses both handcrafted and deep geometric features of 3D point clouds into the RGB feature space, refining the RGB-to-3D mapping process and making the model aware of the 3D geometric structure of industrial objects.


\section{Methodology}
\label{sec3}

In this section, we present our multimodal framework for industrial anomaly detection, which synergistically integrates visual textures, geometric structures, and linguistic priors. Unlike existing methods that treat different modalities as isolated features, our approach focuses on the intrinsic correlation between 2D RGB data and 3D depth information while leveraging text-based object-conditioned knowledge to refine the feature space. The overall architecture consists of two core components designed to handle modality alignment and semantic adaptation. First, the Geometry-Aware Cross-Modal Mapper is introduced to bridge the gap between appearance and structure by explicitly decoupling geometric cues from RGB features, ensuring a high-fidelity mapping to the depth domain. Second, to incorporate high-level categorical guidance, we propose the Object-Conditioned Textual Feature Adaptor, which utilizes a Mixture-of-Experts mechanism and cross-attention to adapt textual embeddings into robust visual-linguistic anchors. By jointly optimizing these modules, our framework achieves a more discriminative and unified representation for precise anomaly localization across diverse industrial scenarios.

\subsection{Geometry-Aware Cross-Modal Mapper}
\label{sec3:gacm}
We propose the Geometry-Aware Cross-Modal Mapper (GACM) to bridge the modality gap between RGB texture and Depth geometry. Conventional mapping functions often struggle to decouple structural information from complex color textures, leading to suboptimal alignment. GACM addresses this by explicitly bifurcating the RGB feature $F_{rgb}$ into semantic and geometric components:
\begin{equation}
F_{sem} = \phi_{s}(F_{rgb}),
\end{equation}
\begin{equation}
F_{geo} = \phi_{g}(F_{rgb}),
\end{equation}
where $\phi_{s}$ denotes the linear embedding layer responsible for extracting semantic and appearance-aware features, and $\phi_{g}$ denotes another linear embedding layer dedicated to learning geometry-oriented structural features.

To suppress redundant semantic noise and emphasize structural cues, we introduce a Geometry Prior Gating (GPG) mechanism. The gating signal is derived from the geometric branch via $\mathcal{G} = \sigma(W_g F_{geo})$, which adaptively calibrates the fusion of features:
\begin{equation}
F_{fused} = F_{geo} \odot \mathcal{G} + F_{sem} \odot (1 - \mathcal{G}).
\end{equation}

The calibrated features are then mapped to the depth domain through a non-linear mimicry network consisting of Layer Normalization and GELU activation. To preserve the fundamental linear correlations and stabilize training, a residual connection $\mathcal{R}(F_{rgb})$ is integrated, yielding the final geometry-aware depth representation $F_{depth}$. This design ensures that the mapped features are strictly constrained by the geometric manifold of the target modality.
\begin{equation}
F_{rgb \to 3d} = \mathcal{R}(F_{rgb}) + \text{GELU}(\text{LN}(F_{fused})),
\end{equation}
where $\mathcal{R}(F_{rgb})$ represents the residual connection from the input RGB features. $\text{LN}(\cdot)$ denotes the Layer Normalization operation. $\text{GELU}(\cdot)$ represents the GELU activation function. $F_{rgb \to 3d}$ represents the mapping from RGB features to 3D features.

To bridge the modality gaps, we employ dual-layer Multi-Layer Perceptrons (MLPs) to perform cross-modal mapping. Specifically, the mapped features for RGB-to-text, 3D-to-text, and 3D-to-RGB transformations are denoted as $F_{rgb \to text}$, $F_{3d \to text}$, and $F_{3d \to rgb}$, respectively, where each is obtained through its corresponding projection layer.

\subsection{Object-Conditioned Textual Feature Adaptor}

We design the Object-Conditioned Textual Feature Adaptor (OCTA) to incorporate high-level linguistic priors into the multi-modal alignment framework. Unlike static text embeddings, OCTA dynamically refines textual descriptors to align with specific object categories using a MoE architecture followed by a Transformer-style refinement block. Given an initial text embedding $T$, we first employ a Top-K MoE module to capture diverse semantic nuances by routing the input through specialized experts:
\begin{equation}
\hat{F}_{text} = \sum_{i=1}^{k} g_i(F_{text}) E_i(F_{text}),
\end{equation}
where $g_i$ and $E_i$ denote the gating network and experts, respectively. This allows the model to select the most relevant linguistic experts based on the object's class.

To translate these enhanced textual priors into a format compatible with visual feature spaces, we introduce a set of learnable prototypes $P \in \mathbb{R}^{1 \times D}$. These prototypes act as the Query ($Q$) in a Cross-Attention mechanism, while the enhanced text $T_{enhanced}$ serves as both the Key ($K$) and Value ($V$):
\begin{equation}
F_{p} = \text{Softmax}\left(\frac{(P W_Q) (\hat{F}_{text} W_K)^T}{\sqrt{d}}\right) (\hat{F}_{text} W_V),
\end{equation}
where $W_i \in \mathbb{R}^{D \times D}$ are learnable projection matrices that map the input features into the Query, Key, and Value spaces, respectively. The term $d$ denotes the scaling factor, which prevents the dot product from growing too large in magnitude, thereby ensuring stable gradients during the Softmax operation.

Finally, to further strengthen the non-linear expression of the adapted features, $F_{p}$ undergoes nonlinear transformation using the following formula:
\begin{eqnarray}
\hat{F}_p &=& \text{MLP}(P+F_p),\\
F_{text \to rgb} &=& F_{text \to 3d}  \nonumber \\
                 &=& \text{LN}(\hat{F}_p + \text{Dropout}(\text{FFN}(\hat{F}_p))).
\end{eqnarray}

By conditioning learnable prototypes on textual semantics through this dual-stage refinement, OCTA effectively bridges the gap between discrete linguistic concepts and continuous visual features, providing a robust textual anchor for multimodal IAD.

The construction of high-level textual priors inspired by state-of-the-art IAD frameworks such as WinCLIP \cite{winclip} and AnomalyGPT \cite{anomalygpt}. Specifically, to provide the model with a robust semantic anchor for normality, we design a comprehensive set of normal state text templates for each object class.

These templates are meticulously crafted to define the defect-free state of an object from multiple linguistic perspectives. The core descriptors include direct class labels [c] and state enhancement terms such as flawless [c], perfect [c], and unblemished [c]. Furthermore, we incorporate explicit exclusionary phrases such as [c] without flaw, [c] without defect, and [c] without damage to reinforce the semantic boundary of a normal sample.

To ensure compatibility with pretrained vision-language models, these state-specific descriptors [s] are embedded into standard contextual prompts, such as "a photo of a [s]." and "a photo of the [s].". For a given object class $c \in \text{classes}$ (e.g., bagel, cookie, cable gland), the resulting collection of normal state sentences is processed through a text tokenizer to generate the initial text embeddings $T$. This combinatorial approach ensures that the model captures a wide spectrum of "normal" semantic variations, providing a dense linguistic foundation for the subsequent alignment process.

The fusion of these diverse templates with specific class information constitutes the Object-Conditioned Text that drives the OCTA module. By pairing class-specific identifiers with multiple "normalcy" descriptors, we create a high dimensional textual prior that is inherently conditioned on the object’s identity. When these embeddings $T$ are fed into the OCTA's MoE architecture, the gating network effectively navigates this multi-template space to distill the most representative linguistic features for that specific class.

Finally, these object-conditioned priors serve as the primary input for the prototype-based cross-attention mechanism. This process effectively bridges the gap between discrete normal state descriptions and continuous visual features, establishing a stable and discriminative textual anchor for multimodal anomaly alignment. By conditioning learnable prototypes on these diverse linguistic combinations, OCTA ensures that the resulting features are both class-aware and specialized for detecting deviations from the established "normal" distribution.

\begin{figure}[t]
\centering
\includegraphics[scale=0.5]{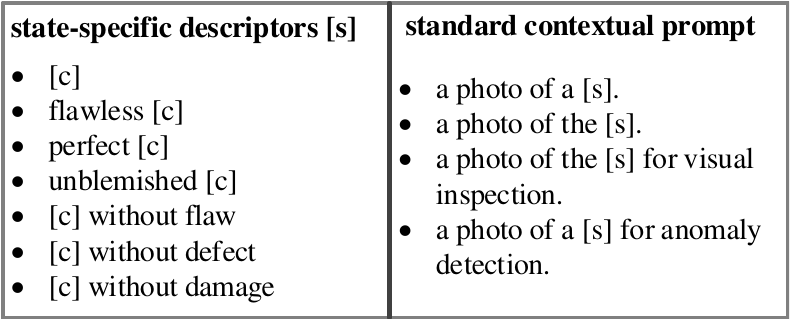}
\caption{Schematic of the Object-Conditioned Textual Prior Generation. $[c]$ Representing different classes. The framework constructs diverse linguistic anchors by embedding state-specific descriptors $[s]$, such as "flawless $[c]$" or "$[c]$ without defect", into standard contextual prompts like "a photo of a $[s]$." to define the normal distribution for each object class $[c]$.}

\label{fig:prompts}
\end{figure}

\subsection{Optimization and Loss Functions}

To achieve a unified and discriminative feature representation, we formulate a comprehensive objective function that facilitates multi-modal consistency. The training process is supervised by two categories of alignment losses: Visual-Geometric Consistency and Visual-Linguistic Alignment.

\subsubsection{Visual-Geometric Consistency}

We employ a bidirectional mapping strategy to enforce the synergy between 2D RGB textural features and 3D geometric structures. Given the predicted 3D features $\hat{F}_{3d}$ from the 2D RGB branch and the predicted 2D RGB features $\hat{F}_{2D}$ from the 3D branch, we minimize the distance between the predicted features and their corresponding groundtruth patches. The loss $\mathcal{L}_{vis}$ is defined as:
\begin{equation}
\begin{aligned}
\mathcal{L}_{vis} = \lambda_{v2g} (1 - \text{sim}(F_{rgb \to 3d}, F_{3d})) + \\
\lambda_{g2v} (1 - \text{sim}(F_{3d \to rgb}, F_{rgb})),
\end{aligned}
\end{equation}
where $\text{sim}(\cdot)$ denotes the cosine similarity metric and $\lambda$ represents the balancing coefficients. Notably, we apply a spatial mask $M_{xyz}$ during the calculation to exclude invalid regions (e.g., background or missing depth pixels), ensuring that the model focuses on the effective object surface. This bidirectional constraint forces the image encoder to internalize geometric awareness while enabling the 3D encoder to retain essential textural context.

\subsubsection{Visual-Linguistic Alignment}

To incorporate high-level semantic priors, we align the visual features from both modalities with the object-conditioned textual embeddings $F_{text \to rgb}$ and $F_{text \to 3d}$ generated by the OCTA module. The textual alignment loss $\mathcal{L}_{text}$ is formulated as:
\begin{equation}
\begin{aligned}
\mathcal{L}_{text} = \lambda_{v2t} (1 - \text{sim}(F_{text \to rgb}, F_{rgb \to text})) + \\
\lambda_{g2t} (1 - \text{sim}(F_{text \to 3d}, F_{3d \to text})),
\end{aligned}
\end{equation}
where $F_{rgb \to text}$ and $F_{3d \to text}$ represents the modality-specific features projected into the textual embedding space. By pulling visual features toward their corresponding linguistic descriptors, the model learns a semantically constrained manifold that is more robust to localized noise and industrial variations.

\subsubsection{Total Objective}

The final training objective is a weighted combination of the aforementioned alignment terms. The total loss $\mathcal{L}_{total}$ is minimized in an end-to-end manner:
\begin{equation}
\mathcal{L}_{total} = \mathcal{L}_{vis} + \mathcal{L}_{text}.
\end{equation}

In our implementation, all balancing hyper-parameters (i.e., $\lambda_{v2g}$, $\lambda_{g2v}$, $\lambda_{v2t}$, and $\lambda_{g2t}$) are empirically set to 1 by default, treating the geometric mapping and linguistic alignment with equal importance. To ensure numerical stability, we implement a validity check to filter out any non-finite gradients before the optimization step. This joint optimization ensures that any discrepancy in the cross-modal consensus—whether arising from textural anomalies or geometric deformations—will manifest as a high anomaly score during inference.

\input{tables/mvtec_3d_full}

\input{tables/mvtec_3d_aupro01}

\subsection{Anomaly Map Generation}
During the inference phase, the matching degrees between the augmented class-specific text features and the RGB/3D features are calculated using Euclidean distance for the current sample class. By integrating these with the cross-modal mapping discrepancies of the original RGB and 3D features, three core criteria for anomaly detection are derived:

\begin{equation}
\Psi_{rgb} = D(F_{rgb}, F_{3d \to rgb} ),
\end{equation}
\begin{equation}
\Psi_{3d} = D(F_{3d}, F_{rgb \to 3d}),
\end{equation}
\begin{equation}
\Psi_{Text} = D(F_{text \to rgb}, F_{rgb \to text}) \odot D(F_{text \to 3d}, F_{3d \to text}),
\end{equation}
\begin{equation}
D(x,y) = \sqrt{\sum_{i=1}^{D}(x_i - y_i)^2},
\end{equation}
where $\odot$ denotes the element-wise multiplication, $\Psi \in \mathbb{R}^{H \times W}$ represents the anomaly map generated during the quality inspection process.$\Psi_{rgb}$ represents the anomaly map calculated as the distance between the original RGB features and the 3D features projected into the 2D RGB space.$\Psi_{3d}$ represents the anomaly map calculated as the distance between the original 3D features and the RGB features projected into the 3D space. $\Psi_{Text}$ denotes the anomaly map generated by measuring the discrepancy between the text embeddings and the aligned multimodal (RGB-3D) features.

Finally, the comprehensive anomaly map $\Psi_{final} \in \mathbb{R}^{H \times W}$ is obtained by fusing the spatial consistency maps with the linguistic alignment map. We employ two hyperparameters, $\alpha$ and $\beta$, to balance the contribution of the cross-modal geometric constraints and the text-based semantic constraints:
\begin{equation}
\Psi_{final} = \alpha (\Psi_{rgb} \odot \Psi_{3d}) + \beta \Psi_{text},
\end{equation}
where $\odot$ denotes the element-wise multiplication. This fusion strategy ensures that the final inspection result benefits from both structural point-to-pixel consistency and high-level semantic verification. By default, we set both $\alpha$ and $\beta$ to 0.5.

\section{Experiments}
\label{sec4}

\subsection{Experimental setups}

\subsubsection{Datasets}
We conduct experiments on two mainstream RGB-3D IAD datasets, namely MVTec 3D-AD \cite{mvtec3d} and Eyecandies \cite{eyecandies}. 
MVTec 3D-AD is a classic real world dataset in the field of 3D industrial anomaly inspection. It covers 10 common industrial object categories, including Bagel, Cable Gland, Carrot, etc. In total, the dataset provides 2,656 training samples and 1,137 test samples. Data acquisition is performed using a structured-light industrial scanner, which simultaneously captures 3D point clouds and corresponding RGB images of target objects. This dataset faithfully reflects various anomaly scenarios in real industrial production, such as surface defects, shape deformations, and other typical manufacturing flaws.

Eyecandies is a synthetic multi-modal anomaly detection dataset proposed for unsupervised visual inspection and anomaly localization. It consists of 10 candy themed object categories, including Candy Cane, Chocolate Cookie and other confectionery items, with a total of 15,000 samples covering 10,000 normal training samples, 1,000 normal validation samples and 4,000 test samples with pixel-level anomaly annotations. All instances are generated via modeling software, and data are captured in a simulated industrial conveyor belt environment. The dataset provides paired RGB images and depth images, covering challenging scenarios including complex textures, self-occlusions, and specular reflections, which effectively validate the generalization ability of the model under diverse environmental conditions.

\subsubsection{Evaluation metrics}
Following the standard evaluation protocol in unsupervised anomaly detection research, we employ several widely recognized metrics to comprehensively evaluate model performance. Among these, the Area Under the Receiver Operating Characteristic curve (AUROC) is a core indicator for measuring the ability to distinguish between normal and anomalous samples. Specifically, we use image-level AUROC (I-AUROC) to quantify the global discrimination capability between normal and anomalous images, and pixel-level AUROC (P-AUROC) to assess the fine-grained ranking performance of anomaly scores across normal and abnormal pixels. Notably, P-AUROC is robust to the severe class imbalance commonly encountered in anomaly detection tasks. In addition, we adopt the Area Under Per-region Overlap curve (AUPRO) to evaluate the spatial coverage quality between predicted anomaly regions and ground-truth defect regions at the component level. We report AUPRO values at two representative thresholds, namely AUPRO@30\% and AUPRO@1\%, to quantitatively analyze localization accuracy under relatively relaxed and strict criteria, respectively.
Collectively, these complementary metrics allow a thorough assessment of our method in terms of image-level discrimination, pixel-level anomaly ranking, and region-level defect localization under the unsupervised setting.



\subsubsection{Implementation details}

We employ the same frozen Transformers as previous studies \cite{m3dm,cfm} to realize the RGB and 3D feature extractors, i.e., DINO ViT-B/8 \cite{DINO,dino_ref2} trained on ImageNet \cite{imagenet} and Point-MAE \cite{pointmae} trained on ShapeNet \cite{shapenet}, respectively. Thus, RGB feature extractor processes 224 × 224 RGB images and outputs 28 × 28 × 768 feature maps, which are bilinearly up-sampled to 224 × 224 × 768 before feeding the features to GACM. 3D feature extractor processes 1024 groups of 32 points obtained with FPS \cite{pointnet++}, yielding a feature vector of dimensionality 1152 for each group. As described in in \ref{sec3:gacm}, these features are interpolated and aligned to 224 × 224 × 1152 before being fed to $\phi_{3d \to rgb}(\cdot)$.

The method we propose is to train the unified model under unsupervised settings, where no manual annotation of class labels is involved during the training process, and the model automatically learns discriminative features across different classes. Meanwhile, to conduct a fair comparison, we reproduce the training of "one-model-one-class" methods under the same unified setting, such as CFM \cite{cfm} and G2SF \cite{g2sf}. 

\input{tables/eyecandies_auroc}

\input{tables/eyecandies_aupro}

\subsection{Main results}

\subsubsection{Quantitative comparison}
The performance of the proposed method is evaluated against several state-of-the-art baselines on the MVTec 3D-AD dataset, with the results summarized in Table \ref{tab:mvtec}. Quantitative comparisons are conducted using two primary metrics: I\_AUROC and AUPRO at a 30\% integration limit. As indicated by the empirical data, the proposed approach consistently outperforms existing methods across the majority of object categories, achieving a superior average performance of 93.99\% in I\_AUROC and 96.99\% in AUPRO. This represents a significant margin over the second best performing method, CFM, which yields an average I\_AUROC of 91.58. The robustness of the proposed framework is particularly evident in challenging categories such as cookie and bagel, where it achieves near-perfect I\_AUROC scores of 99.55\% and 99.33\%, respectively.

A comparative analysis of individual classes reveals that while baseline models exhibit significant performance fluctuations, the proposed method maintains high stability. For instance, in the cookie class, competitive models such as MambaAD and G2SF struggle significantly, recording I\_AUROC values of only 61.17\% and 78.32\%. In contrast, the proposed method effectively addresses these difficult cases, demonstrating its superior ability to capture complex 3D geometric anomalies and subtle surface defects. Although MambaAD shows a slight advantage in the cable gland class in terms of I\_AUROC, the proposed method provides a more balanced trade-off, as evidenced by its higher AUPRO scores and overall consistency across the entire dataset. These results collectively validate that the integration of our proposed modules leads to more precise anomaly localization and classification, setting a new benchmark for 3D industrial defect detection.

\input{tables/ablation_module}

Table \ref{tab:mvtec_aupro01} summarizes the anomaly localization performance on the MVTec 3D-AD dataset using the AUPRO@1\% metric. Our proposed method achieves an average score of 44.87\%, outperforming state-of-the-art methods such as CFM and M3DM. Specifically, our approach demonstrates significant improvements in challenging categories like cookie (+4.85\% over CFM) and rope (+3.54\% over CFM). The consistent superiority across most classes validates the effectiveness of our method.

As shown in Table \ref{tab:eyecandies_auroc}, our proposed method demonstrates a clear performance advantage over existing baselines on the Eyecandies dataset, achieving a superior average I-AUROC of 86.19\% and P-AUROC of 97.76\%. This significant margin—surpassing the next best method, CFM, by 8.91\% in image-level detection—is particularly evident in challenging categories such as ChocolateCookie and Confetto, where our approach exceeds baseline results by over 16\%. Furthermore, the consistency of our model is underscored by its near perfect pixel-level localization across almost all candy types, such as achieving good performance with 99.64\% P-AUROC in marshmallow, effectively validating that our multimodal feature alignment strategy provides a more robust and granular representation for identifying anomalies than current state-of-the-art architectures.

In terms of the anomaly localization performance presented in Table \ref{tab:eyecandies_aupro}, our method significantly outperforms the baseline models across both integration thresholds, achieving a leading average AUPRO@30\% of 89.60\% and AUPRO@1\% of 34.48\%. The performance gap is particularly pronounced in the more stringent AUPRO@1\% metric, where our approach exceeds the state-of-the-art CFM by 5.6\% and more than doubles the accuracy of G2SF. This substantial improvement is consistently observed across challenging categories such as ChocolateCookie and LicoriceSandwich, where our method improves upon the best baseline by 8.06\% and 8.07\% respectively, demonstrating that our multimodal mapping effectively captures fine-grained structural anomalies. It is worth noting that memory-bank-based methods like G2SF face inherent challenges in the unified anomaly detection setting; the requirement to store an extensive volume of normal samples across all categories not only imposes a significant memory overhead but also risks feature interference between classes, where overlapping cross classes representations can lead to increased false positives and degraded localization precision. In contrast, our approach sidesteps these bottlenecks, proving more robust in distinguishing subtle defects across heterogeneous candy types.

\subsubsection{Qualitative comparison}
\begin{figure}[t]
\centering
\includegraphics[scale=0.4]{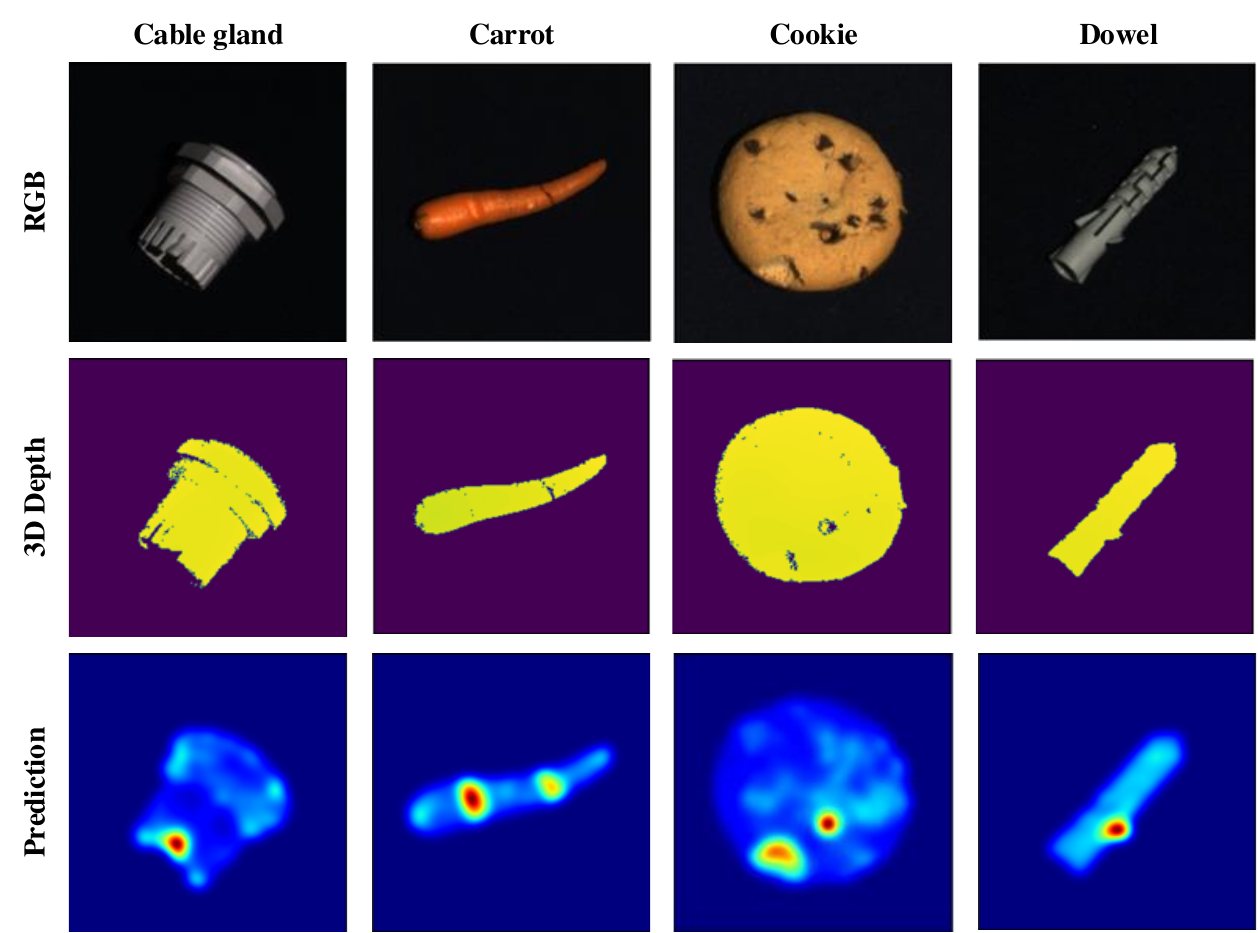}
\caption{
  Qualitative Results. From top to bottom: input RGB images, 3D Dpeth images, and anomaly maps with our method.
  }

\label{fig:visual}
\end{figure}

As illustrated in Fig. \ref{fig:visual}, we present a qualitative comparison of anomaly localization results. Our approach generates remarkably sharper and more precise anomaly maps that align closely with the groundtruth defect segmentations. While baseline methods often produce blurred responses or suffer from background noise, particularly in unified scenarios where multi-class features can interfere our model maintains high structural fidelity and localized focus. This superior visualization capability directly motivates the significant performance gains observed in our quantitative evaluation, especially regarding the rigorous AUPRO@1\% metric.

\subsection{Ablation study}
In this section, we conduct extensive ablation studies and qualitative analyses to evaluate the effectiveness of the proposed components and their individual contributions to the overall performance. 
We first perform a component-wise investigation to validate the necessity of each module in multimodal industrial anomaly detection. 
Furthermore, we assess the robustness of our framework through a sensitivity analysis of key hyperparameters. 
Finally, we investigate the influence of different prompt engineering strategies to demonstrate how semantic guidance impacts the reasoning process and detection accuracy.

\begin{figure}[t]
\centering
\includegraphics[scale=0.35]{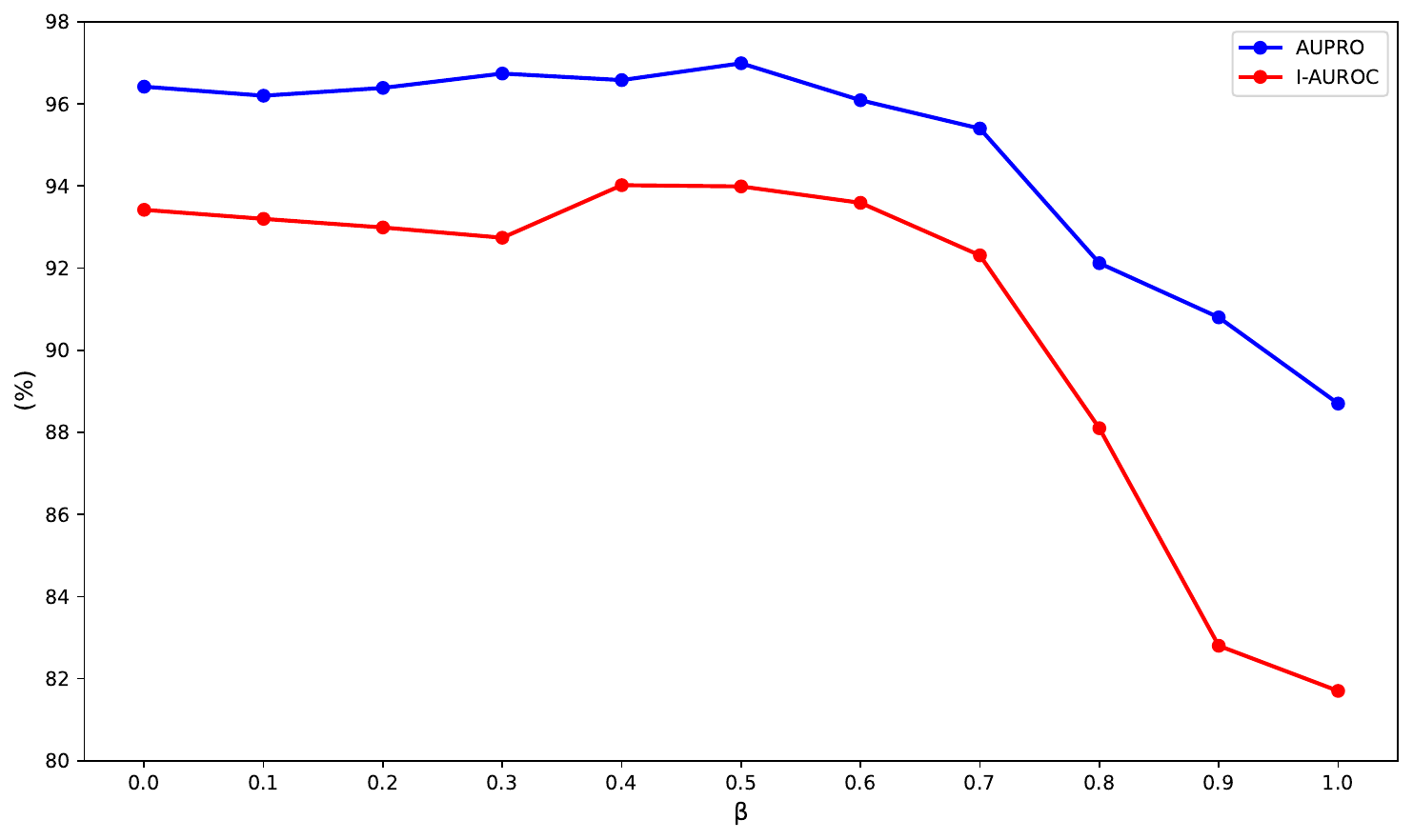}
\caption{The impact of $\beta$.}
\label{fig:beta}
\end{figure}

\subsubsection{Influence of different components}
To evaluate the individual and collective contributions of the Geometry-Aware Cross-Modal Mapper, the Object-Conditioned Textual Feature Adaptor, and Layers Pruning, we conducted an ablation study on the MVTec 3D-AD dataset, as summarized in Table \ref{tab:ablation}. Starting from the baseline, the introduction of either GACM or OCTA in conjunction with Layers Pruning leads to a marked improvement in image-level detection. While the baseline exhibits competitive performance in pixel-level metrics, the synergistic integration of all three components yields the most robust results across the board. Notably, the full model achieves the highest localization precision, recorded at 99.34\% P-AUROC and 44.87\% AUPRO@1\%. These results demonstrate that while OCTA significantly bolsters global anomaly recognition, the addition of GACM and Layers Pruning is essential for refining fine-grained anomaly localization, ultimately confirming that the three modules complement each other to bridge the modality gap and filter redundant features effectively.


\subsubsection{Hyperparameter Sensitivity Analysis}

In this section, we evaluate the impact of the fusion coefficients $\alpha$ and $\beta$ on the final anomaly detection performance. These hyperparameters control the relative contribution of the cross-modal spatial consistency $(\Psi_{rgb} \odot \Psi_{3d})$ and the text-based semantic verification $(\Psi_{text})$. By default, we set $\alpha = 0.5$ and $\beta = 0.5$ to balance the geometric and linguistic constraints. To investigate their sensitivity, we conducted comparative experiments by varying their values (e.g., $0.4$ and $0.6$). As shown in \ref{fig:beta}, the model achieves the most robust performance when $\alpha$ and $\beta$ are balanced, while deviating from this equilibrium leads to a slight decrease in AUPRO@1\%, suggesting that both structural consistency and semantic alignment are equally indispensable for high-precision anomaly localization.

\subsubsection{Influence of prompts}

To evaluate the impact of our proposed prompt engineering and training strategies, we conduct an ablation study as presented in Table \ref{tab:ablation_prompts}. The results highlight the performance gains achieved by Object-Conditioned Text, Compositional Prompt Ensemble (CPE) \cite{winclip}, and the Abnormal Prompt Generator (APG) \cite{iadgpt}. Notably, while the baseline achieves a respectable I-AUROC of 91.58\%, the introduction of individual components leads to substantial improvements across all metrics. Both CPE and APG are integrated into a contrastive learning framework for self-supervised training, where synthetic data generation is implemented following the approach of EasyNet to simulate realistic anomaly patterns. Within this contrastive paradigm, CPE provides a diverse semantic representation of normal states, while APG synthesizes targeted negative samples to refine the decision boundary between normal and anomalous features. The results show that APG alone pushes the AUPRO@1\% to 44.74\%, underscoring its ability to generate effective supervisory signals. However, the optimal performance is reached through the OCT configuration, which achieves a peak I-AUROC of 94.02\% and a P-AUROC of 99.35\%.

\input{tables/ablation_prompts}

\subsubsection{Efficiency analysis}

As illustrated in Table~\ref{tab:efficiency_comparison}, we conduct a comprehensive evaluation of efficiency and detection performance. It is worth noting that M3DM operates under a "one-model-one-class" setting, which inherently simplifies the task but leads to massive cumulative memory overhead and limited scalability. In contrast, our method, along with CFM and G2SF, follows the more challenging unified multi-class setting.

Under this unified paradigm, our method demonstrates significant superiority. While CFM achieves the lowest memory footprint, its detection accuracy (I-AUROC) and fine-grained localization (AUPRO@1\%) lag behind ours. Compared to G2SF, our approach achieves a nearly 12 times reduction in memory usage and a 1.7 times boost in inference speed, all while maintaining substantially higher precision across all metrics. Remarkably, even when compared to the class-specific M3DM, our unified model achieves highly competitive I-AUROC and superior pixel-level localization. These results underscore that our method strikes an optimal balance between high efficiency and robust generalization, making it exceptionally well-suited for real-time, large-scale industrial anomaly detection.


\section{Conclusion}
\label{sec5}

This work presents a unified multimodal framework for unsupervised RGB‑3D industrial anomaly detection, addressing the core limitations of existing methods: insufficient semantic guidance, weak geometric modeling, and restrictive single‑class training. By integrating text semantic priors, a geometry‑aware cross‑modal mapper, and class‑unified representation learning, our framework aligns RGB, 3D and text features in a shared semantic space while preserving fine‑grained geometric structure. Experiments on MVTec 3D‑AD and Eyecandies show that our approach achieves state‑of‑the‑art performance in unified detection and segmentation. Ablation studies verify the effectiveness of each module. This work offers a practical solution for multi‑class, high‑precision visual inspection.

\input{tables/ablation_mem_fps}

In the future, we can further explore the extension of the proposed multimodal framework to zero-shot and few-shot industrial anomaly detection scenarios to enhance its generalization to unseen defects and new product categories with limited labeled data.

\ifCLASSOPTIONcaptionsoff
  \newpage
\fi

\bibliographystyle{IEEEtran}
\bibliography{IEEEabrv,reference}


\end{document}

%% file: tables/mvtec_3d_full.tex
\begin{table*}[tbp]
\centering
\setlength{\tabcolsep}{3pt} 
\caption{\MakeUppercase{
Performance comparison of different methods in MVTec 3D-AD. The values in the table are presented as I\_AUROC/AUPRO@30\%. \textbf{Bold} indicates the best results.}}
\vspace{5pt}
\label{tab:mvtec}
\begin{tabular}{lcccccc}
\toprule
Classes & MambaAD \cite{mambaad} & CFM \cite{cfm} & BTF \cite{btf} & G2SF \cite{g2sf} & Ours \\
\midrule
bagel        & 87.71/92.07 & 97.52/\textbf{97.75} & 91.80/97.60 & 94.96/96.59 & \textbf{99.33}/\textbf{97.99} \\
cable gland  & \textbf{94.31}/\textbf{98.40} & 83.36/95.51 & 74.80/96.90 & 73.78/87.81 & 83.09/95.41 \\
carrot       & 90.69/98.15 & 96.66/\textbf{98.21} & 96.70/97.90 & 93.22/97.49 & \textbf{97.00}/98.17 \\
cookie       & 61.17/83.65 & 99.51/93.96 & 88.30/\textbf{97.30} & 78.32/90.15 & \textbf{99.55}/95.16 \\
dowel        & 97.60/\textbf{97.05} & 94.97/91.82 & 93.20/93.30 & 78.41/88.29 & \textbf{98.82}/95.27 \\
foam         & 84.00/82.73 & 86.06/\textbf{96.38} & 58.20/88.80 & 74.81/83.51 & \textbf{86.88}/96.18 \\
peach        & 92.78/97.06 & 91.18/97.89 & 89.60/97.50 & 86.71/95.50 & \textbf{92.85}/\textbf{98.00} \\
potato       & 66.85/94.77 & 89.08/\textbf{98.22} & 91.20/98.10 & 87.49/97.24 & \textbf{91.55}/98.00 \\
rope         & 97.37/95.52 & 94.93/96.63 & 92.10/95.00 & 87.98/84.39 & \textbf{98.96}/\textbf{97.65} \\
tire         & 90.02/97.04 & 82.48/97.88 & 88.60/97.10 & 64.00/77.86 & \textbf{91.86}/\textbf{98.11} \\
\midrule
Average      & 86.25/93.64 & 91.58/96.42 & 86.50/95.90 & 81.97/89.88 & \textbf{93.99}/\textbf{96.99} \\
\bottomrule
\end{tabular}
\end{table*}

%% file: tables/mvtec_3d_aupro01.tex
\begin{table*}[tbp]
\centering
\setlength{\tabcolsep}{5pt}  
\caption{\MakeUppercase{Performance of different methods on MVTec 3D-AD dataset (aupro@1\%).  \textbf{Bold} indicates the best results.}}
\label{tab:mvtec_aupro01}
\begin{tabular}{lccccccc}
\toprule
Classes & BTF \cite{btf} & AST \cite{ast} & M3DM \cite{m3dm} & CFM \cite{cfm} & G2SF \cite{g2sf} & Ours \\
\midrule
bagel            & 42.80        & 38.80        & 41.40         & 44.66        & 41.87         & \textbf{46.46}         \\
cable gland     & 36.50        & 32.20        & 39.50         & 37.82        & 23.89         & \textbf{36.80}         \\
carrot           & 45.20        & 47.00        & 44.70         & \textbf{48.68}        & 45.51         & 48.18         \\
cookie           & 43.10        & 41.10        & 31.80         & 45.83        & 30.61         & \textbf{50.68}         \\
dowel            & 37.00        & 32.80        & \textbf{42.20}         & 33.28        & 28.65         & 39.42         \\
foam             & 24.40        & 27.50        & 33.50         & \textbf{42.14}        & 29.43         & 40.73         \\
peach            & 42.70        & \textbf{47.40}        & 44.40         & 45.19        & 36.53         & 46.26         \\
potato           & 47.00        & \textbf{48.70}        & 35.10         & 48.37        & 40.01         & 46.15         \\
rope             & 29.80        & 36.00        & 41.60         & 43.21        & 30.60         & \textbf{46.75}         \\
tire             & 34.50        & 47.40        & 39.80         & 46.27        & 18.61         & \textbf{47.26}         \\
\midrule
Average             & 38.30        & 39.80        & 39.40         & 43.54        & 32.57         & \textbf{44.87}         \\
\bottomrule
\end{tabular}
\end{table*}

%% file: tables/eyecandies_auroc.tex
\begin{table*}[tbph]
\centering
\small
\setlength{\tabcolsep}{4pt}
\caption{\MakeUppercase{Anomaly detection performance in terms of Image-level AUROC (I-AUROC) and Pixel-level AUROC (P-AUROC) on the Eyecandies dataset. \textbf{Bold} indicates the best results.}}
\label{tab:eyecandies_auroc}
\begin{tabular}{l|ccc|ccc}
\toprule
\multirow{2}{*}{Classes} & \multicolumn{3}{c|}{I-AUROC} & \multicolumn{3}{c}{P-AUROC} \\
\cmidrule(lr){2-7}
& CFM \cite{cfm} & G2SF \cite{g2sf} & Ours & CFM \cite{cfm} & G2SF \cite{g2sf} & Ours \\
\midrule
CandyCane         & 47.84 & \textbf{51.68} & 48.48 & 97.91 & 97.07 & \textbf{98.27} \\
ChocolateCookie   & 76.48 & 66.96 & \textbf{93.12} & 96.81 & 92.39 & \textbf{98.70} \\
ChocolatePraline  & 88.16 & 80.91 & \textbf{92.64} & 95.13 & 91.69 & \textbf{96.83} \\
Confetto          & 76.80 & 76.85 & \textbf{95.20} & 98.32 & 93.89 & \textbf{99.57} \\
GummyBear         & 88.14 & 80.98 & \textbf{91.03} & 95.05 & 91.74 & \textbf{95.59} \\
HazelnutTruffle   & \textbf{80.64} & 52.13 & 74.88 & 95.09 & 87.37 & \textbf{94.09} \\
LicoriceSandwich  & 73.44 & 71.76 & \textbf{84.96} & 95.64 & 92.11 & \textbf{97.26} \\
Lollipop          & 66.73 & 71.64 & \textbf{87.50} & 97.68 & 97.73 & \textbf{98.41} \\
Marshmallow       & 96.96 & 75.36 & \textbf{98.56} & 99.25 & 91.62 & \textbf{99.64} \\
PeppermintCandy   & 77.60 & 81.42 & \textbf{95.52} & 98.59 & 93.68 & \textbf{99.19} \\
\midrule
\textbf{Average}     & 77.28 & 70.97 & \textbf{86.19} & 96.95 & 92.93 & \textbf{97.76} \\
\bottomrule
\end{tabular}
\end{table*}

%% file: tables/eyecandies_aupro.tex
\begin{table*}[tbph]
\centering
\small
\setlength{\tabcolsep}{4pt}
\caption{\MakeUppercase{Anomaly localization performance (AUPRO) at 30\% and 1\% thresholds on the Eyecandies dataset. \textbf{Bold} indicates the best results.}}
\label{tab:eyecandies_aupro}
\begin{tabular}{l|ccc|ccc}
\toprule
\multirow{2}{*}{Classes} & \multicolumn{3}{c|}{AUPRO @30\%} & \multicolumn{3}{c}{AUPRO @1\%} \\
\cmidrule(lr){2-4} \cmidrule(lr){5-7}
& CFM \cite{cfm} & G2SF \cite{g2sf} & Ours & CFM \cite{cfm} & G2SF \cite{g2sf} & Ours \\
\midrule
CandyCane         & 92.91 & 90.34 & \textbf{94.57} & 15.26 & 13.20 & \textbf{21.41}  \\
ChocolateCookie   & 85.21 & 71.97 & \textbf{91.88} & 31.94 & 15.73 & \textbf{40.00}  \\
ChocolatePraline  & 76.84 & 68.24 & \textbf{81.63} & 29.60 & 17.23 & \textbf{36.10}  \\
Confetto          & 96.31 & 78.86 & \textbf{97.37} & 40.51 & 19.44 & \textbf{43.80}  \\
GummyBear         & 85.46 & 79.70 & \textbf{86.81} & 32.43 & 19.99 & \textbf{33.76}  \\
HazelnutTruffle   & 72.85 & 57.68 & \textbf{79.19} & 16.37 & 6.72 & \textbf{20.21}  \\
LicoriceSandwich  & 78.63 & 68.64 & \textbf{82.22} & 25.45 & 12.21 & \textbf{33.52}  \\
Lollipop          & 88.30 & 88.40 & \textbf{91.34} & 14.25 & 17.47 & \textbf{29.08}  \\
Marshmallow       & 93.92 & 73.11 & \textbf{95.21} & 44.05 & 20.43 & \textbf{46.25}  \\
PeppermintCandy   & 94.86 & 81.26 & \textbf{95.72} & 38.97 & 21.91 & \textbf{40.70}  \\
\midrule
\textbf{Average}     & 86.53 & 75.82 & \textbf{89.60} & 28.88 & 16.43 & \textbf{34.48}  \\
\bottomrule
\end{tabular}
\end{table*}

%% file: tables/ablation_module.tex

\begin{table}[tbph]
\centering
\small
\setlength{\tabcolsep}{4pt}
\caption{\MakeUppercase{Ablation study of different components on the MVTec 3D-AD dataset. GACM, OCTA, and LP represent Geometry-Aware Cross-Modal Mapper, Object-Conditioned Textual Feature Adaptor, and Layers Pruning respectively. \textbf{Bold} indicates the best performance.}}
\label{tab:ablation}
\begin{tabular}{ccc|cc}
\toprule
\multicolumn{3}{c|}{Components} & \multicolumn{2}{c}{Metrics} \\
GACM & OCFA & Layers Pruning & I-AUROC & AUPRO@30\% \\
\midrule
             &              & $\checkmark$ & 91.58 & 96.42 \\
$\checkmark$ &              & $\checkmark$ & 93.47 & 96.72 \\
             & $\checkmark$ & $\checkmark$ & \textbf{94.02} & 96.81 \\
$\checkmark$ & $\checkmark$ &              & 93.78 & 96.92 \\
$\checkmark$ & $\checkmark$ & $\checkmark$ & 93.99  & \textbf{96.99} \\
\bottomrule
\end{tabular}
\end{table}

%% file: tables/ablation_prompts.tex

\begin{table}[t]
\centering
\small
\setlength{\tabcolsep}{4pt} 
\caption{\MakeUppercase{Ablation study of different prompt-related components on the MVTec 3D-AD dataset. OCT, CPE, and APG represent Object-Conditioned Text, Compositional Prompt Ensemble \cite{winclip}, and Abnormal Prompt Generator \cite{iadgpt}, respectively. \textbf{Bold} indicates the best performance.}}
\label{tab:ablation_prompts}
\begin{tabular}{ccc|cc}
\toprule
\multicolumn{3}{c|}{Components} & \multicolumn{2}{c}{Metrics} \\
OCT & CPE & APG & I-AUROC & AUPRO@30\% \\
\midrule
             &              &              & 91.58 & 96.42 \\
$\checkmark$ &              &              & \textbf{94.02} & \textbf{97.12} \\
             & $\checkmark$ &              & 93.57 & 96.95 \\
             &              & $\checkmark$ & 93.90 & 96.99 \\
\bottomrule
\end{tabular}
\end{table}

%% file: tables/ablation_mem_fps.tex

\begin{table}[t]
\centering
\small
\setlength{\tabcolsep}{3pt} 
\caption{\MakeUppercase{Comparison of memory footprint, inference speed, and performance on MVTec 3D-AD. \textbf{Bold} indicates the best performance among models under the unified multi-class setting.}}
\label{tab:efficiency_comparison}
\begin{tabular}{l|cc|cc}
\toprule
Method & Memory (MB) $\downarrow$ & FPS $\uparrow$ & I-AUROC & AUPRO@30\% \\
\midrule
M3DM & 11962.0 & 0.7 & \textbf{94.5} & 96.4 \\
CFM  & \textbf{621.0} & \textbf{21.8} & 91.6 & 96.4 \\
G2SF & 12803.4 & 6.0 & 82.0 & 89.9 \\
\textbf{Ours} & 1089.8 & 10.1 & 94.0 & \textbf{97.0} \\
\bottomrule
\end{tabular}
\end{table}